\documentclass[conference,a4paper]{IEEEtran}
\IEEEoverridecommandlockouts
\usepackage[numbers]{natbib}
\usepackage{tablefootnote}
\usepackage[T1]{fontenc}
\usepackage{booktabs}
\usepackage{amsmath,amssymb,amsfonts}
\usepackage{algorithmic}
\usepackage{graphicx}
\usepackage{textcomp}
\usepackage{microtype}
\usepackage{bm}
\usepackage{enumitem}
\usepackage{chngcntr}
\usepackage{algorithm}
\usepackage[dvipsnames]{xcolor}
\usepackage{hyperref}
\newcommand\tab[1][2em]{\hspace*{#1}}
\def\tablename{Table}
\def\BibTeX{{\rm B\kern-.05em{\sc i\kern-.025em b}\kern-.08em
    T\kern-.1667em\lower.7ex\hbox{E}\kern-.125emX}}

\newcommand{\forceindent}{\leavevmode{\parindent=1em\indent}}

\usepackage{fancyhdr}
\usepackage{lastpage}
\usepackage{afterpage}
\begin{document}

\title{Enhancing Deep Learning-based 3-lead ECG Classification with Heartbeat Counting and Demographic Data Integration
}

\author{
\IEEEauthorblockN{
    Khiem H. Le $^{a,b,*}$, Hieu H. Pham $^{a,b}$, Thao BT. Nguyen $^{a,b}$, Tu A. Nguyen $^{a,b}$, Cuong D. Do $^{a,b}$ \thanks{* Corresponding author: \url{khiem.lh@vinuni.edu.vn}}
}
\IEEEauthorblockA{$^{a}$College of Engineering and Computer Science, VinUniversity, Hanoi, Vietnam \\ $^{b}$VinUni-Illinois Smart Health Center, VinUniversity, Hanoi, Vietnam
}
}

\maketitle

\forceindent
\begin{abstract}
Nowadays, an increasing number of people are being diagnosed with cardiovascular diseases (CVDs), the leading cause of death globally. The gold standard for identifying these heart problems is via electrocardiogram (ECG). The standard 12-lead ECG is widely used in clinical practice and the majority of current research. However, using a lower number of leads can make ECG more pervasive as it can be integrated with portable or wearable devices. This article introduces two novel techniques to improve the performance of the current deep learning system for 3-lead ECG classification, making it comparable with models that are trained using standard 12-lead ECG. Specifically, we propose a multi-task learning scheme in the form of the number of heartbeats regression and an effective mechanism to integrate patient demographic data into the system. With these two advancements, we got classification performance in terms of F1 scores of 0.9796 and 0.8140 on two large-scale ECG datasets, i.e., Chapman and CPSC-2018, respectively, which surpassed current state-of-the-art ECG classification methods, even those trained on 12-lead data. To encourage further development, our source code is publicly available at \url{https://github.com/lhkhiem28/LightX3ECG}. 
\end{abstract}


\section{Introduction}
\begin{figure}[ht]
    \centering
    \includegraphics[keepaspectratio, height=225.7pt]{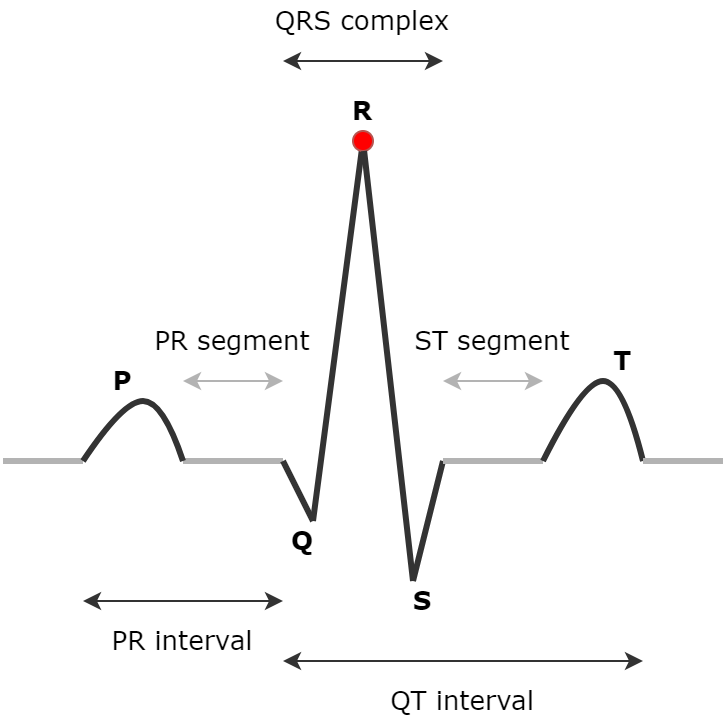}
    \caption{The usual structure of an ECG beat \cite{our}.}
    \label{fig:ECG}
\end{figure}

\forceindent
Cardiovascular diseases (CVDs) are a group of disorders of the heart and blood vessels, including coronary heart disease, cerebrovascular disease, rheumatic heart disease, and many other conditions. CVDs are one of the primary causes of death globally, accounting for approximately 19.05 million deaths worldwide in 2020 \cite{Statistics-2022}. Fortunately, premature deaths can be avoided by early detection of people who are most at risk for CVDs and ensuring they receive the appropriate care with counseling and medications. The electrocardiogram (ECG), which is a representation of the electrical activity of the heart obtained by placing electrodes on the body surface, shows the pathological states of the cardiovascular system in its waveforms or rhythms. Therefore, this electrical signal is a widely used, simple but effective tool to identify cardiovascular abnormalities in patients. The usual structure of an ECG beat is shown in Fig \ref{fig:ECG}, consisting of a P wave, a QRS complex with an R peak, a T wave, and other parts such as PR, QT intervals, or PR, ST segments. 
\\\\
It is estimated that 1.5 million and 3 million electrocardiograms are conducted every single day across the globe \cite{ECG-test-annual-1, ECG-test-annual-2}, this makes computer-aided, automatic ECG analysis essential, especially in low- and middle-income countries, where high-quality and experienced cardiologists are extremely scarce. In the past few years, research works concerning automatic ECG analysis have witnessed rapid development, and deep learning-based methods have become the preferred approach. For standard 12-lead ECG, Acharya et al. \cite{9_layer_CNN_5_classes} early developed a 9-layer 1D-CNN to identify 5 different types of cardiovascular abnormalities, Zhang et al. \cite{GradientSHAP} proposed using 1D-ResNet34, Zhu et al. \cite{34_layer_CNN_27_classes} ensembled two 1D-SEResNet34s and one set of expert rules to identify more types of abnormalities. Li et al. \cite{Sleep_Apnea} combined a sparse Autoencoder and Hidden Markov Model for diagnosing obstructive sleep apnea. Moreover, Rahman et al. \cite{COV-ECGNET} tried to early diagnose COVID-19 using ECG trace images. Recently, with the development of many small, low-cost, and easy-to-use ECG-enable devices \cite{Zio_Patch, Digital_Stethoscope, Apple_Watch} which provide a subset of ECG leads rather than the entire set, newer methods are being developed to do ECG classification based on reduced-lead ECG rather than standard 12-lead data. For instance, Drew et al. \cite{myocardial_ischemia} demonstrated that interpolated 12-lead ECG, which is derived from a reduced-lead set (limb leads plus V1 and V5), is comparable to standard 12-lead ECG for diagnosing wide-QRS-complex tachycardias and acute myocardial ischemia. Green et al. \cite{coronary_syndrome} also found that the leads III, aVL, and V2 together yielded a similar performance as the full 12-lead ECG for diagnosing acute coronary syndrome. Cho et al. \cite{myocardial_infarction} claimed that myocardial infarction could be detected not only with a conventional 12-lead ECG but also with a limb 6-lead ECG. Unlike previous studies, our work uses a combination of only three ECG leads (I, II, and V1) as the input for the system to strike a balance between high classification performance and the ease of signal acquisition, thus providing further support to demonstrate the ability of reduced-lead ECG to identify a wider range of cardiovascular abnormalities. 
\\\\
From the perspective of deep learning, there are two primary strategies to enhance the performance of a learning-based model, that are internal improvement and external information integration. For internal improvements, existing research works mainly focus on model architecture changes or robust loss function usages. For instance, Cheng et al. \cite{ECG-signal-deep_CNN-and-BiLSTM} used a much larger kernel size in order to capture longer patterns in ECG signals. Yao et al. \cite{TI-CNN} constructed a Time-Incremental ResNet18 (TI-ResNet18), a combination of a 1D-ResNet18 and an LSTM network to capture both spatial and temporal features. Zhu et al. \cite{34_layer_CNN_27_classes} suggested using a Sign loss function and Romdhane et al. \cite{focal_loss} suggested using a Focal loss function to reduce the negative effects caused by class imbalance problems in datasets. In this work, we propose a novel, simple but effective yet internal improvement as a multi-task learning scheme where we perform a subtask of the number of heartbeats regression. In general, a multi-task learning scheme offers the advantage of reducing overfitting through shared representations, thus improving the system's generalization. Moreover, by performing the proposed heartbeat counting subtask, the system is relatively forced to capture periodic features in ECG signals, which are mostly ignored in other methods. 
\\\\
The idea of utilizing external data or metadata to improve classification performance has been explored in other domains. In particular, Li et al. \cite{GPS-info} attempted to enhance landmark identification in tourist photographs by using GPS information. Ellen et al. \cite{plankton} incorporated both geo-temporal and hydrographic metadata to boost the performance of plankton image classification. In the medical domain, Ningrum et al. \cite{Melanoma} observed the use of patient metadata such as age, gender, and anatomical site leads to higher accuracy in malignant melanoma detection on dermoscopic images. In this work, we design an effective encoding process and a shallow neural network to embed patient demographic data, including age and gender, which along with ECG recordings, to further increase the power of the developed ECG classification system. 
\\\\
To summarize, our main contributions are as follows:
\begin{itemize}[leftmargin=*]
\item We propose a novel multi-task learning scheme in the form of the number of heartbeats regression to help the system capture periodic features and increase its generalization
\item An effective mechanism is designed to integrate patient demographic data, including age and gender, into the system, thus boosting the classification performance
\item Extensive experiments were conducted on two large-scale multi-lead ECG datasets, i.e., Chapman and CPSC-2018, where the enhanced system significantly outperformed state-of-the-art models that are trained using 12-lead ECG
\end{itemize}

\section{Baseline System}
\forceindent
In this section, we briefly introduce our current system, X3ECG \cite{our}. Firstly, the overall architecture is described. Next, we present the most important part, the Lead-wise Attention module, in more detail. 

\subsection{Overall Architecture}
We employ three distinct 1D-SEResNets18 \cite{Squeeze-and-Excitation} as backbones to extract features from three input ECG leads separately. This multi-input strategy is reasonable since these kinds of signals usually require separate treatment. Then, a novel Lead-wise Attention module is designed as the aggregation technique to explore the most essential input lead and merge outputs of these backbones, resulting in a more robust representation that is then sent through a Fully-Connected (FC) layer to perform classification. 

\subsection{Lead-wise Attention}
\begin{figure}[!ht]
    \centering
    \includegraphics[keepaspectratio, width=244.4pt]{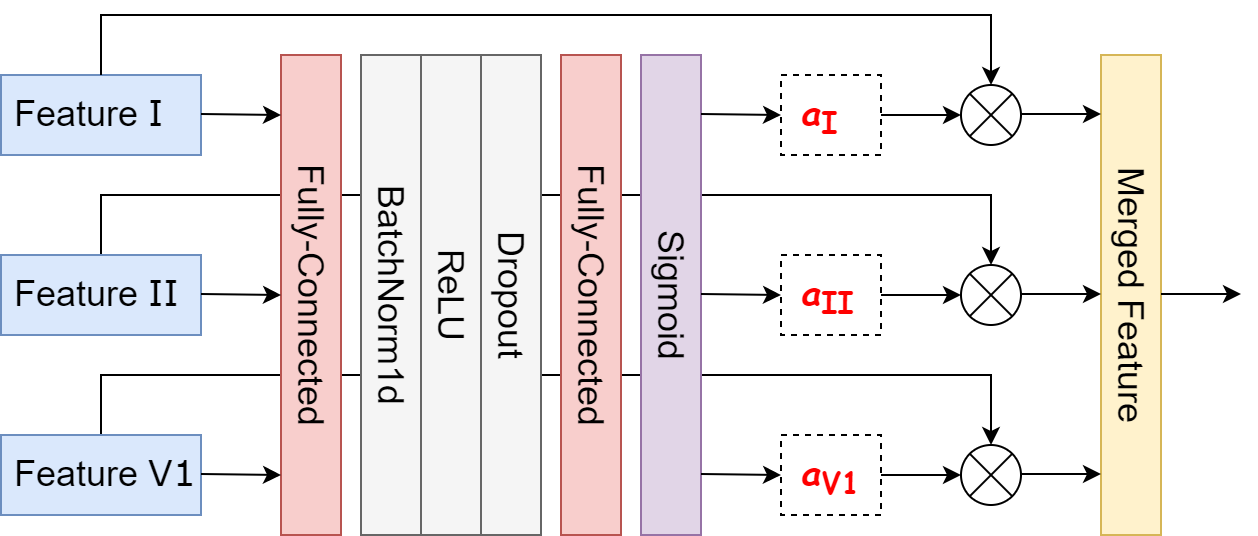}
    \caption{The proposed Lead-wise Attention module.}
    \label{fig:Lead-wise Attention}
\end{figure}

To achieve an end-to-end classification system, the outputs, also known as features, extracted from backbones, must be combined. Typically, one can combine these features by simply applying a summation or concatenation operation to them, but this is usually ineffective due to their simplicity. Thus, we propose a Lead-wise Attention module to more effectively ensemble these features together and acquire a final robust feature which is then routed to the last FC layer, the classifier, to perform classification. Our Lead-wise Attention module is described in Fig \ref{fig:Lead-wise Attention}. 
\\\\
Firstly, features from our backbones are concatenated and sent through a sequential list of layers including an FC, a BatchNorm, a Dropout, followed by another FC layer and a Sigmoid function to determine the attention score, or importance score for each feature. Subsequently, the final feature is obtained by taking a weighted sum over these features by corresponding generated scores. This module can be formulated as: 
\begin{gather*}
\text{f}_{\text{merged}}=\sum_{i=1}^{3}\boldsymbol{\alpha}_{i}\text{f}_{i}, 
\\
\boldsymbol{\alpha}=\texttt{Sigmoid}(\texttt{FC}(\texttt{FC}(\texttt{Concat}[\text{f}_{i}|i=\overline{1,3}]))).
\end{gather*}
The BatchNorm, Dropout layer, and ReLU function are ignored in the above formulation for simplicity. 

\section{Proposed Techniques}
\begin{figure*}
    \centering
    \includegraphics[keepaspectratio, width=480pt]{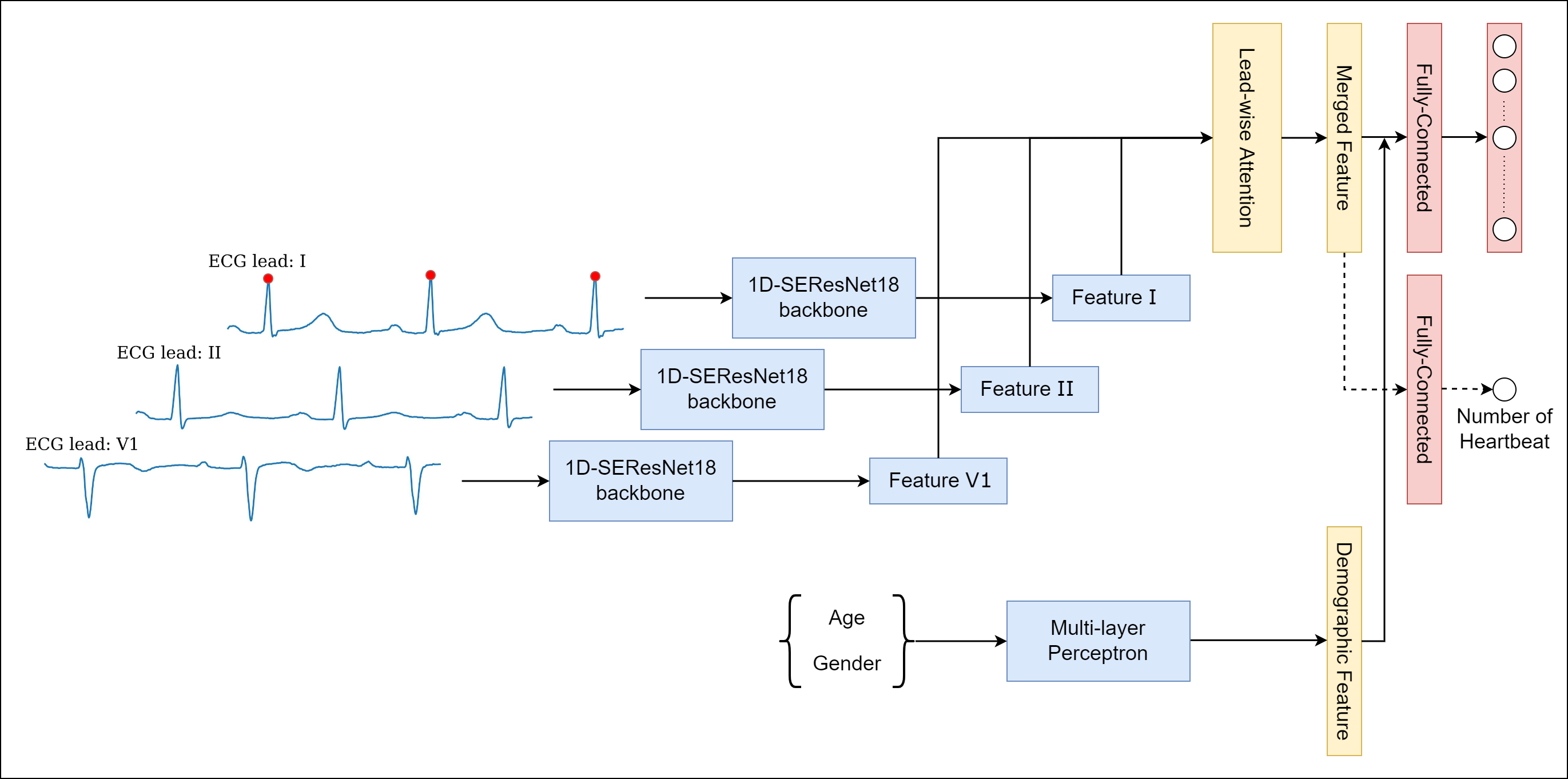}
    \caption{An overview of the whole proposed system.}
    \label{fig:System}
\end{figure*}

\forceindent
In this section, we sequentially describe the proposed internal improvement, namely heartbeat counting (HC) and the mechanism of demographic data integration (DDI), that form the final enhanced system, as illustrated in Fig \ref{fig:System}. 

\subsection{Heartbeat Counting}
We formulate the heartbeat counting subtask as a simple regression problem. Specifically, an FC layer with one output unit is used as an auxiliary regressor, taking the merged feature from the Lead-wise Attention module $\text{f}_{\text{merged}}$ as input and trying to predict how many heartbeats there are in ECG recording: 
\begin{gather*}
\text{n}_{pred}=\texttt{FC}(\text{f}_{\text{merged}})
\end{gather*}
At the training stage, the number of heartbeats of each ECG recording is required as ground truth. But the majority of datasets lack this information. Alternatively, we utilize a popular and highly accurate tool called NeuroKit2 \cite{NeuroKit2} to detect R peaks in lead I ECG signal of a recording. The number of R peaks represents the number of heartbeats, thus is considered ground truth. A Mean Absolute Error (MAE) is employed to compute the difference between ground truth $\text{n}_{gt}$ and the predicted number of heartbeats $\text{n}_{pred}$. Finally, the whole system is trained end-to-end with a combined loss function as: 
\begin{gather*}
\mathcal{L}_{combined}=\mathcal{L}_{cls} + \lambda|\text{n}_{gt} - \text{n}_{pred}|,
\end{gather*}
where $\mathcal{L}_{cls}$ is the main CVDs classification loss function, and $\lambda$ is a hyperparameter, which controls the contribution and effect of the auxiliary regressor on the whole system. At the inference stage, the auxiliary branch will be ignored, thus there is no additional cost to the system. 

\subsection{Demographic Data Integration}
Different research has demonstrated that demographic data, including age and gender, is associated with cardiovascular risks. Also, cardiologists need to additionally use this information to make final decisions \cite{Rodgers2019-xe, Jousilahti1999-nf}. Based on that, we leverage this external data, available in most ECG datasets to enhance the system's performance further. 

\subsubsection{Feature Encoding}\hfill\\
We have age and gender features in the form of discrete and categorical features, respectively, that need an effective encoding strategy before feeding into the system. Firstly, the age feature is converted into a categorical feature by being divided into 7 groups including 0 to 4 years, 5 to 14 years, 15 to 22 years, 23 to 41 years, 42 to 56 years, 57 to 68 years, and over 69 years that are respectively corresponding to early childhood, childhood, adolescence, youth, maturity, old age, and late old age. Then, we apply one-hot encoding to these two categorical features and treat missing values as their own category. Finally, we obtain a one-hot feature vector with a length of 11, which can be used by the system. 
\\
\subsubsection{Feature Fusion}\hfill\\
To achieve an end-to-end system, we design a shallow neural network, also known as Multi-layer Perceptron (MLP) to process the encoded demographic data and fuse the output feature with the merged feature from the Lead-wise Attention module $\text{f}_{\text{merged}}$ by a simple concatenation operation. 
\begin{algorithm}
\caption{PyTorch-like pseudocode for the Feature Fusion}\label{alg:mlp}
\tab $\texttt{MLP = nn.Sequential(}$\\
\tab \tab $\texttt{nn.Linear(\textcolor{Green}{11}, \textcolor{Green}{128}), }$\\
\tab \tab $\texttt{nn.BatchNorm1d(\textcolor{Green}{128}), }$\\
\tab \tab $\texttt{nn.ReLU(), }$\\
\tab \tab $\texttt{nn.Dropout(\textcolor{Green}{0.2}), }$\\
\tab \tab $\texttt{nn.Linear(\textcolor{Green}{128}, \textcolor{Green}{128}), }$\\
\tab \tab $\texttt{nn.BatchNorm1d(\textcolor{Green}{128}), }$\\
\tab \tab $\texttt{nn.ReLU(), }$\\
\tab $\texttt{)}$\\
\tab $\texttt{f\_demog = MLP(demog)}$\\
\tab $\texttt{f\_final = torch.\textcolor{Violet}{cat}([f\_demog,f\_merged])}$
\end{algorithm}
\\
Algorithm \ref{alg:mlp} provides the pseudo-code for our designed MLP architecture and the fusion operation in a PyTorch-like style \cite{PyTorch}. The final feature is used for performing the main task CVDs classification. 

\section{Experiments and Results}
\forceindent
In this section, we evaluate the proposed techniques on two large-scale ECG datasets, i.e., Chapman and CPSC-2018, with class frequency and demographic data statistics shown in Table \ref{tab:datasets}. Extensive experiments were performed to compare our system with other ECG classification methods and to validate the contributions of each component in the whole system. 

\subsection{Datasets}
\textit{Chapman} \cite{Chapman}. Chapman University and Shaoxing People's Hospital collaborated to establish this large-scale \textit{multi-class} ECG dataset which consisted of 10.646 12-lead ECG recordings. Each recording was taken over 10 seconds with a sampling rate of 500 Hz and labeled with 11 common diagnostic classes. The amplitude unit was microvolt. These 11 classes were hierarchically grouped into 4 categories including AFIB, GSVT, SB, and SR. 
\begin{table}[ht]
\normalsize
\setlength{\tabcolsep}{5.5pt}
\renewcommand{\arraystretch}{1}
\caption{Description of two datasets.}
\begin{tabular}{lrrc}
\midrule
\multicolumn{4}{l}{Chapman} \\
\midrule
Class & Frequency (\%) & Male (\%) & Age \\
\midrule
AFIB & 2225 (20.90) & 1298 (58.34) & 72.90 $\pm$ 11.68 \\
GSVT & 2307 (21.67) & 1152 (49.93) & 55.44 $\pm$ 20.49 \\
SB   & 3889 (36.53) & 2481 (63.80) & 58.34 $\pm$ 13.95 \\
SR   & 2225 (20.90) & 1025 (46.07) & 50.84 $\pm$ 19.25 \\
\midrule
\midrule
\multicolumn{4}{l}{CPSC-2018} \\
\midrule
Class & Frequency (\%) & Male (\%) & Age \\
\midrule
SRN  & 918  (13.35) & 363  (39.54) & 41.56 $\pm$ 18.45 \\
AF   & 1221 (17.75) & 692  (56.67) & 71.47 $\pm$ 12.53 \\
IAVB & 722  (10.50) & 490  (67.87) & 66.97 $\pm$ 15.67 \\
LBBB & 236  (03.43) & 117  (49.58) & 70.48 $\pm$ 12.55 \\
RBBB & 1857 (27.00) & 1203 (64.78) & 62.84 $\pm$ 17.07 \\
PAC  & 616  (08.96) & 328  (53.25) & 66.56 $\pm$ 17.71 \\
PVC  & 700  (10.18) & 357  (51.00) & 58.37 $\pm$ 17.90 \\
STD  & 869  (12.64) & 252  (29.00) & 54.61 $\pm$ 17.49 \\
STE  & 220  (03.20) & 180  (81.82) & 52.32 $\pm$ 19.77 \\
\midrule
\end{tabular}
\label{tab:datasets}
\end{table}
\\
\textit{CPSC-2018} \cite{CPSC-2018}. The first China Physiological Signal Challenge was held in 2018, and a large-scale \textit{multi-label} ECG dataset was made accessible to the public. This dataset contained 6.877 12-lead ECG recordings with a sampling rate of 500 Hz, an amplitude unit of millivolt, and durations ranging from 6 to 60 seconds. These ECG recordings were labeled with 9 diagnostic classes including SNR, AF, IAVB, LBBB, RBBB, PAC, PVC, STD, and STE. 

\subsection{Implementation Details}
As a deep learning system requires inputs to be of the same length, all ECG recordings were fixed at 10 seconds in length in both datasets. This was done by truncating the part exceeding the first 10 seconds for longer recordings and padding shorter ones with zero. Then, signals were filtered using a zero-phase method with 3rd order Butterworth bandpass filter with a frequency band from 1 Hz to 47 Hz. We took leads I, II, and V1 from each ECG recording to construct the input with the shape of 3x5000 and fed it into the system. 
\\\\
For evaluation, we applied a 10-fold cross-validation strategy following some previous works \cite{GradientSHAP, Chapman}. We stratify divided each of the two datasets into 10 folds and performed 10 rounds of training and evaluation. At each round, 8 folds; 1 fold; and 1 remaining fold were used as training, validation, and test set, respectively. In the multi-label classification manner, the optimal threshold of each class was searched in a range (0.05, 0.95) with a step of 0.05 to achieve the best F1 score on the validation set. We report the average performance of 10 rounds on the test set in terms of F1 score and accuracy. In all experiments on our system, the system was optimized by an Adam optimizer with an initial learning rate of 1e-3 and a weight decay of 5e-5 for 70 epochs. We used the Cosine Annealing scheduler in the first 40 epochs to reschedule the learning rate to 1e-4 and then kept it constant in the last 30 epochs. Cross-entropy and binary cross-entropy were utilized as classification loss functions in multi-class and multi-label manners, respectively. When applying the proposed heartbeat counting technique, the control hyperparameter $\lambda$ was set to 0.02. All experiments were performed on a workstation with an NVIDIA GeForce RTX 3090 TURBO 24G. 

\begin{table*}[!t]
\normalsize
\setlength{\tabcolsep}{6pt}
\renewcommand{\arraystretch}{1.06}
\caption{Comparison of the whole proposed system to SOTA methods.}
\begin{tabular}{l|rrrr}
\midrule
Method & Chapman macro-F1 & Chapman Acc & CPSC-2018 macro-F1 & CPSC-2018 Acc \\
\midrule
TI-ResNet18 \cite{TI-CNN} & 0.9348 & 0.9399 & 0.7387 & 0.9484 \\
InceptionTime \cite{InceptionTime} & 0.9640 & 0.9671 & 0.7908 & 0.9604 \\
1D-SEResNet34 \cite{GradientSHAP} & 0.9695 & 0.9723 & 0.8036 & 0.9635 \\
1D-SEResNet34 w/ Focal loss \cite{focal_loss} & 0.9722 & 0.9738 & 0.8095 & 0.9640 \\
X3ECG \cite{our} & 0.9746 & 0.9775 & 0.8025 & 0.9637 \\
X3ECG w/ HC & 0.9783 & 0.9803 & 0.8110 & 0.9648 \\
X3ECG w/ HC + DDI & \textbf{0.9796} & \textbf{0.9817} & \textbf{0.8140} & \textbf{0.9652} \\
\midrule
\end{tabular}
\label{tab:comparison}
\end{table*}

\subsection{Comparison with SOTA Methods}
We compared the whole proposed system with state-of-the-art ECG classification methods including TI-ResNet18 \cite{TI-CNN}, InceptionTime \cite{InceptionTime}, 1D-SEResNet34 \cite{GradientSHAP}, and 1D-SEResNet34 with Focal loss \cite{focal_loss}. While our system was trained on only three ECG leads, all other methods were trained using standard 12-lead ECG. From Table \ref{tab:comparison}, X3ECG enhanced with the HC technique helped us to set new state-of-the-art performances on two datasets, outperforming other methods in terms of F1 score and accuracy. Furthermore, the system enhanced with both two techniques, HC and DDI, boosted these performances further. In particular, we got F1 scores of 0.9796 and 0.8140, and accuracies of 0.9817 and 0.9652 on two datasets, Chapman and CPSC-2018 respectively. 

\section{Conclusion}
\forceindent
In this article, we introduced an accurate deep learning system that uses a combination of three ECG leads (I, II, and V1) to identify cardiovascular abnormalities. Besides that, two novel techniques heartbeat counting and demographic data integration, which were developed to enhance system performance were presented. The whole proposed system surpassed current state-of-the-art ECG classification methods even those trained on standard 12-lead ECG. Also, we emphasize that heartbeat counting and demographic data integration are plug-in techniques and can be easily adapted to other different models that can accelerate other research. 

\bibliography{refs}
\bibliographystyle{unsrt}

\end{document}